\documentclass[conference]{IEEEtran}

\usepackage[nolist]{acronym}
\usepackage[utf8]{inputenc}
\usepackage[T1]{fontenc}
\usepackage{microtype}
\usepackage{booktabs}
\usepackage{url}

\usepackage[keeplastbox,ancient]{flushend}

\usepackage{amsmath}
\usepackage{amssymb}
\usepackage{graphicx}
\usepackage{tabularx}
\usepackage{hyperref} 
\usepackage{cleveref}
\usepackage{subfigure} 
\usepackage{xcolor} 

\definecolor{clr_energy}{RGB}{57,117,121}
\definecolor{clr_entertainment}{RGB}{186,171,155}
\definecolor{clr_health}{RGB}{136,104,156}
\definecolor{clr_safety}{RGB}{137,28,86}
\definecolor{clr_other}{RGB}{212,129,59}

\newcommand{\crowdre}{Crowd RE} 
\newcommand{\blockcomment}[1]{}

\makeatletter

\def\ps@IEEEtitlepagestyle{%
  \def\@oddfoot{\mycopyrightnotice}%
  \def\@evenfoot{}%
}
\def\mycopyrightnotice{%
  {\begin{minipage}{\textwidth}
  \footnotesize \copyright 2020 IEEE. Personal use of this material is permitted. Permission from IEEE must be obtained for all other uses, in any current or future media, including reprinting\slash republishing this material for advertising or promotional purposes, creating new collective works, for resale or redistribution to servers or lists, or reuse of any copyrighted component of this work in other works.
  \end{minipage}
  }
  \gdef\mycopyrightnotice{}
}

\begin{document}

\begin{acronym}[]
	\acro{LDA}{Latent Dirichlet Allocation}
	\acro{CBOW}{Continuous Bag-of-Words}
	\acro{TMD}{Topic Mover's Distance}
	\acro{WMD}{Word Mover's Distance}
\end{acronym}

\title{Topic Modeling on User Stories using Word Mover's Distance}

\author{
\IEEEauthorblockN{Kim Julian G{\"u}lle, Nicholas Ford}
\IEEEauthorblockA{Technische Universit{\"a}t Berlin\\
Berlin, Germany\\
\{k.guelle, nicholas.ford\}@campus.tu-berlin.de}
\and
\IEEEauthorblockN{Patrick Ebel, Florian Brokhausen, Andreas Vogelsang}
\IEEEauthorblockA{Technische Universit{\"a}t Berlin\\
Berlin, Germany\\
\{patrick.ebel, florian.brokhausen, andreas.vogelsang\}@tu-berlin.de}
}

\maketitle


\begin{abstract}
Requirements elicitation has recently been complemented with crowd-based techniques, which continuously involve large, heterogeneous groups of users who express their feedback through a variety of media. Crowd-based elicitation has great potential for engaging with (potential) users early on but also results in large sets of raw and unstructured feedback. Consolidating and analyzing this feedback is a key challenge for turning it into sensible user requirements. 
In this paper, we focus on topic modeling as a means to identify topics within a large set of crowd-generated user stories and compare three approaches: (1) a traditional approach based on Latent Dirichlet Allocation, (2) a combination of word embeddings and principal component analysis, and (3) a combination of word embeddings and Word Mover's Distance.
We evaluate the approaches on a publicly available set of 2,966 user stories written and categorized by crowd workers. We found that a combination of word embeddings and Word Mover's Distance is most promising.
Depending on the word embeddings we use in our approaches, we manage to cluster the user stories in two ways: one that is closer to the original categorization and another that allows new insights into the dataset, e.g.\ to find potentially new categories. Unfortunately, no measure exists to rate the quality of our results objectively. Still, our findings provide a basis for future work towards analyzing crowd-sourced user stories.

\end{abstract}



\section{Introduction} 
\label{sec:introduction}
In traditional Requirements Engineering (RE), techniques like surveys, workshops, observations, and interviews are used to gather stakeholder input and elicit software requirements~\cite{Zowghi2005}. 
Usually, these techniques are limited and can only be applied to end-users within organizational reach~\cite{oriol_fame_2018}. With the emergence of new data sources, this changes: Researchers have shown different approaches to extract requirements from feedback channels such as tweets or app store reviews~\cite{oriol_fame_2018,stanik_classifying_2019}. Another approach to gather a broad range of feedback is crowd-sourcing~\cite{Groen17}. In 2016, Murukannaiah~et~al.\ elicited 2,966 requirements for smart home applications from crowd workers~\cite{murukannaiah_acquiring_2016}. 
Such forms of user feedback can be used to identify, prioritize, and manage requirements~\cite{Maalej16} for software products and to increase user satisfaction~\cite{palomba_user_2015}. 
However, automatic techniques are necessary to derive useful insights the from large amounts of raw data the crowd can produce~\cite{Maalej16,murukannaiah_toward_2017,Groen17}. This becomes even more apparent as the decision-making process in requirements engineering shifts towards a more data-driven approach~\cite{maalej_data-driven_2019}. The automatic analysis of crowd-based requirements comes with some challenges, though, as Murukannaiah~et~al.\ declared~\cite{murukannaiah_toward_2017}. With our paper, we work on how to summarize crowd-acquired requirements automatically. In our evaluation, these requirements are represented in the form of user stories, which seem to be an appropriate form for crowd-sourced requirements elicitation~\cite{Menkveld19,Kolpondinos2019}. Our contribution is a method to cluster requirements through the combined use of topic modeling techniques and similarity metrics based on word embeddings. The goal is to provide the basis for an automatic solution that identifies groups of requirements or features in crowd-sourced data. Existing work for automatic requirements clustering is mostly based on \ac{LDA}, a statistical model that characterizes a requirement by a distribution over certain latent \emph{topics}. In contrast, we cluster user stories based on word embeddings and distance measures. Although an objective and quantified evaluation is not possible in our study setup due to a missing \emph{ground truth}, we conclude that a clustering approach based on pretrained word embeddings and \ac{WMD} as distance measure produced the most promising and interesting results in our setting.

\section{Background}


\subsection{The CrowdRE Dataset}

With the intention to ``facilitate large scale user participation in RE'', Murukannaiah~et~al.~\cite{murukannaiah_toward_2017, murukannaiah_acquiring_2016} conducted an empirical study on the Amazon Mechanical Turk\footnote{\url{https://www.mturk.com/}} platform, resulting in the CrowdRE dataset consisting of 2,966 crowd-generated requirements for smart home applications.

The dataset was generated in two phases: First, 300 crowd workers were asked to formulate requirements for smart home applications in the form of user stories (As a [role], I want [feature] so that [benefit]). The authors had to assign one of five domains to the requirement (\emph{Energy}, \emph{Entertainment}, \emph{Health}, \emph{Safety}, or \emph{Other}). Additionally, an arbitrary number of free-text tags could be added.

In the second phase, 309 additional crowd workers rated the requirements of the first phase with regard to clarity, usefulness, and novelty. For our work, the results of the first phase are the primary data source since we want to extract topics based on the textual data. Below, an exemplary requirement and its annotated domain and tags is given:

\noindent
\begin{tabularx}{\columnwidth}{@{}lX@{}}
Requirement: & \emph{``As a pet owner, I want my smart home to let me know when the dog uses the doggy door, so that I can keep track of the pets whereabouts.''}\\
Domain: & \emph{Safety}\\
Tags: & \emph{Pets, Cats, Dogs}\\
\end{tabularx}



\subsection{Latent Dirichlet Allocation (LDA)} 
\label{sub:lda}

\ac{LDA} proposed by Blei~et~al.~\cite{blei_latent_2003} is a generative probabilistic model used to observe hidden groups of similar data called \emph{topics} within a dataset.
The authors define a word as an item from a vocabulary, a document as a sequence of words, and a corpus as a collection of documents. The approach aims to find a limited number of topics that are latent in the documents of the corpus. To do so, it is assumed that each document is a mixture of a limited number of latent topics with each topic being modeled as the probability distribution over all words in the vocabulary. 
Based on this generative model for a collection of documents, the \ac{LDA} approach uses backtracking to find a set of topics that likely have generated the corpus.
Therefore, for a new document, it is possible to infer the involved latent topics and assign a topic label~\cite{niu_topic2vec_2015}. However, \ac{LDA} suffers from order effects~\cite{Agrawal.2018} i.e.\ if the input data is shuffled, different topics can be retrieved. This leads to different results each time the algorithm computes the topics and therefore introduces new challenges for subsequent text mining algorithms. Additionally, being a probabilistic model, \ac{LDA} models describe the relationship between words as a statistical relationship of occurrences without considering the semantic information embedded in words~\cite{niu_topic2vec_2015}. Therefore, the similarity between words based on their meaning cannot be discovered~\cite{mikolov_efficient_2013} which, in turn, can result in too broad topics~\cite{niu_topic2vec_2015}.


\subsection{Word Vectors and Word Embeddings}
\label{sub:back_word_embeddings}

To overcome the introduced shortcomings, continuous space neural network language models can be trained to capture both the syntactic and the semantic regularities of language. A common defining feature of such models is that each word is converted into a high-dimensional real-valued vector (\textit{word vector}) via learned lookup-tables. A property of these models is that similar words are likely to have similar vectors~\cite{mikolov_linguistic_2013}. 


%
\subsubsection{Word2Vec} 
\label{sub:word_2_vec}

Although several architectures for the computation of word vectors exist~\cite{mikolov_efficient_2013,mikolov_linguistic_2013}, according to Mikolov~et~al., none of these \textit{``architectures has been successfully trained on more than a few hundred of millions of words''}~\cite{mikolov_efficient_2013}, as they become computationally very expensive with larger data sets. This also applies to the previously mentioned LDA. Addressing this shortcoming, Mikolov~et~al.\ propose two optimized neural network architectures for calculating word vectors at a significantly reduced learning time: the \emph{ \ac{CBOW} model} and the \emph{continuous skip-gram model}~\cite{mikolov_efficient_2013}.

The idea behind the \ac{CBOW} architecture is to predict the current word based on the context, whereas the Skip-gram model predicts surrounding words given the current word~\cite{mikolov_efficient_2013}. Both are shallow neural network architectures consisting of an input layer, a projection layer, and an output layer~\cite{mikolov_efficient_2013,qiang_topic_2016}. Once the language model is trained on any of these architectures, the projection layer holds a dense representation of the word vectors, also called \textit{word embedding}\footnote{\url{https://www.tensorflow.org/tutorials/text/word_embeddings}}. These embeddings preserve the syntactic and semantic information of the words. Therefore, when displayed in vector space, it is possible, to express these syntactic and semantic similarities by vector offsets, where all pairs of words sharing a particular relation are related by the same constant offset~\cite{mikolov_linguistic_2013}.


\subsubsection{Word Mover's Distance} 
\label{sub:word_movers_distance}
While word2vec is a sophisticated approach when it comes to generating quality word embeddings, the word vectors alone are not sufficient regarding the task of topic modeling. Consider the two documents: \emph{``My smart home should turn on my favorite music when I come to my home.''} and \emph{``My smart home shall play my most favored songs when I arrive at my place.''} The sentences basically convey the same information. Plotting these sentences with word embeddings, some of their vectors will even be close, especially if word-wise similarity is given (e.g.\ the pairs \emph{\textless music, songs\textgreater} and \emph{\textless come, arrive\textgreater} are close. The closeness of the whole sentences, on the other hand, cannot be represented in the word2vec model alone. To overcome this shortage, Kusner~et~al.\ introduced \acf{WMD} as a word-based distance measure for whole sentences~\cite{kusner_word_2015}. Based on previously created word embeddings (as for example those from word2vec), the distance between two text documents A and B is described as the minimum cumulative distance that words from document A need to travel to match exactly the point cloud of document B. Using this method, \ac{WMD} reaches a high retrieval accuracy while being completely free of hyper-parameters and therefore straightforward to use.


\section{Related Work} 
\label{sec:related_work}

When it comes to topic modeling, the \ac{LDA} algorithm is the most widely-used technique in recent approaches throughout software engineering~\cite{Agrawal.2018}. In the following, we will introduce some \ac{LDA}-based approaches and approaches based on word embeddings.

Guzman and Maalej~\cite{Guzman14} applied NLP and sentiment analysis to extract software features from user reviews together with a summary of the user opinions about each feature. To identify high-level features, they used \ac{LDA} on a set of features they extracted from the reviews.

Galvis Carre{\~n}o and Winbladh~\cite{Carreno13} extracted word-based topics from reviews and assigned sentiments to them through a combination of \ac{LDA} and sentiment analysis. Similarly, Chen~et~al.~\cite{Chen14} proposed AR-miner, a review analytics framework for summarizing informative app reviews. The tool first filters noisy and irrelevant reviews, such as ratings. Then, it summarizes and ranks the informative reviews using topic modeling (\ac{LDA} and Aspect and Sentiment Unification Model (ASUM)) and heuristics from the review metadata.

Another approach is presented by Asuncion~et~al.~\cite{Asuncion.2010} who propose a method that automatically records traceability links and then performs topic modeling. The topic model is learned over the artifacts and allows a semantic categorization and topical visualization of the system. The presented tools aid users to analyze the semantic nature of artifacts and the software architecture itself.

Another \ac{LDA}-based approach is presented by Barua~et~al.~\cite{Barua.2014}. They use \ac{LDA} to automatically identify the main topics in the textual content of Stack Overflow discussions. They additionally quantify how these topics change over time to retrieve emerging trends and gain more detailed insights into the needs of developers. A similar approach is presented by Zhou~et~al.~\cite{zhou_tong_text_2016}. They evaluated over 200,000 Wikipedia articles and as a second analysis applied their \ac{LDA}-based approach to a set of twitter messages from 10,000 users. They were able to retrieve articles as well as twitter users that cover similar content. However, due to the large amount of data, the \ac{LDA}-approach turned out to be computationally expensive.

While the above-mentioned approaches mainly focus on user-generated content (user reviews, Stack Overflow posts), Hindle~et~al.\ have applied \ac{LDA} to extract topics from documented requirements at Microsoft~\cite{Hindle12} and found that many topics were relevant to features and development effort. Stakeholders who were familiar with the requirements documents tended to be comfortable labeling the topics and identifying behavior, but those who were not showed some resistance to the task of topic labeling. 

In contrast to the above-introduced approaches, methods based on recent neural probabilistic language models~\cite{Bengio.2003} have shown that they are able to address the shortcomings introduced by the \ac{LDA}-based approaches. In particular, the already introduced Word2Vec approach proposed by Mikolov~et.~al~\cite{mikolov_efficient_2013} is used by multiple other approaches to build upon. One of these approaches is presented by Qiang~et~al.~\cite{qiang_topic_2016} who propose an embedding-based Topic Model (ETM) that uses semantic knowledge from word embeddings to alleviate the problem of very limited word co-occurrence information in short texts. They claim that their method outperforms the state-of-the-art methods, including LDA, on two real-world datasets.

Another approach, built upon word embeddings, that aims to solve the problem of sparseness in terms of word co-occurrences, is presented by Li~et.~al.~\cite{Li.2016}. They propose a method that is particularly suited to perform topic modeling on short texts by incorporating knowledge about word semantics, learned from a large number of external documents. 

Wu~and~Li~\cite{wu_topic_2017} present an approach called \ac{TMD}, being a topic-based distance metric for documents, inspired by \ac{WMD}. In their approach, each document is considered to be composed of predefined topics and each topic is denoted by a word cluster. These word clusters are then expanded to a vector space in which \ac{TMD} measures how far topics need to travel from one document to another.


\section{Proposed Approaches} 
\label{sec:own_approach}

\begin{figure}
  \centering
    \includegraphics[width=.8\linewidth]{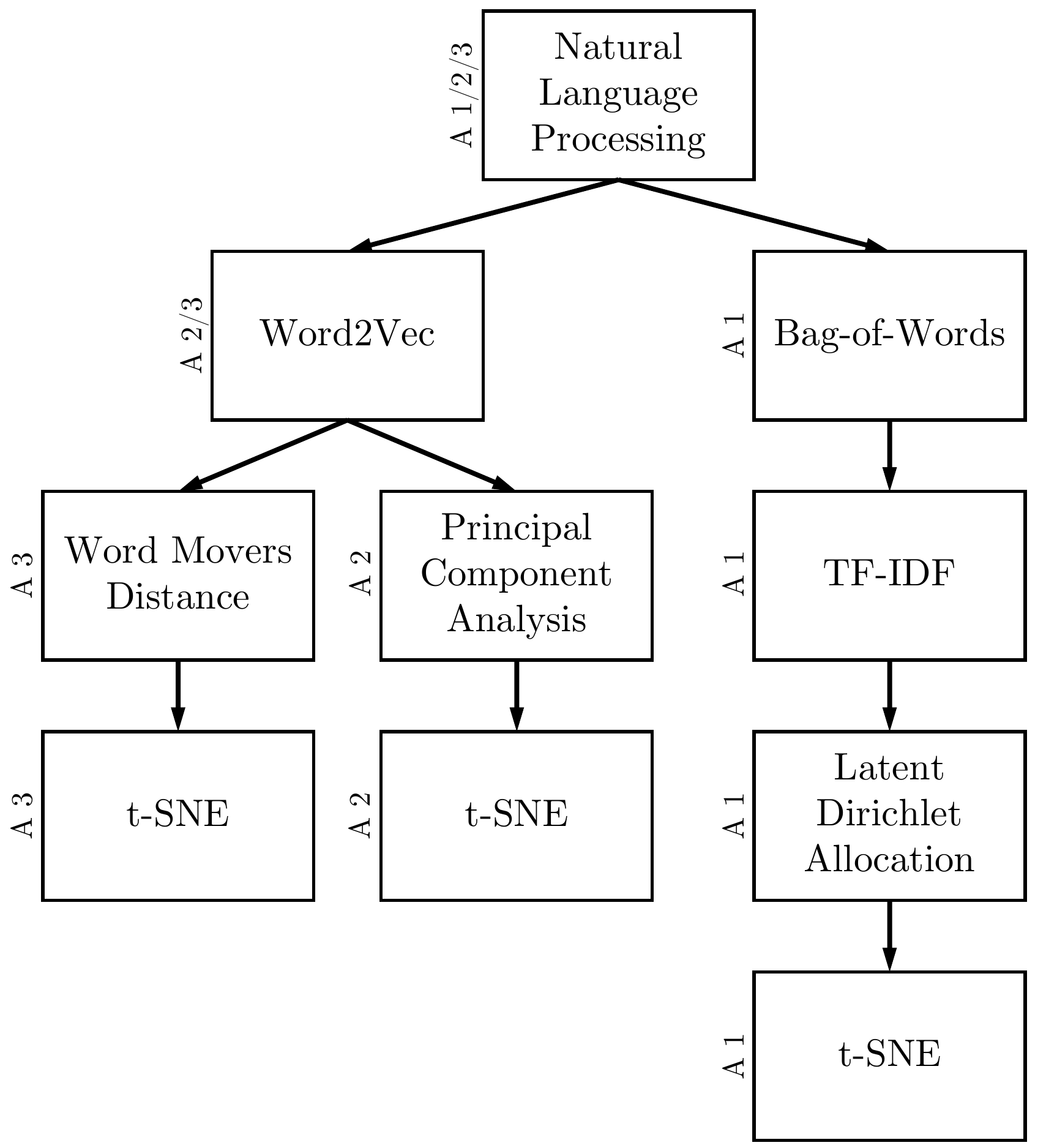}
    \caption{Overview of the analyzed approaches (A1, A2, and A3)}
    \label{fig:approaches_overview}
\end{figure}

In this section, we present the three topic modeling approaches we use. \Cref{fig:approaches_overview} shows an overview of the different process steps for each approach, with \emph{A1 - 3} signifying the three approaches as detailed later in the respective chapters. We published the code of the three approaches to enable replication and reuse.\footnote{\url{https://github.com/firstdayofjune/aire-20}}
Before detailing the separate approaches, some common processing steps are covered.

\subsection{Dataset Preparation}\label{sub:data_preparation}

For our topic modeling approaches, we only use the text of the requirements without any ratings or user characterization added to the data. We therefore extract the sentences of the \crowdre{} dataset\footnote{\url{https://crowdre.github.io/murukannaiah-smarthome-requirements-dataset/}} to construct our dataset.

\begin{figure}
  \centering
    \includegraphics[width=\linewidth]{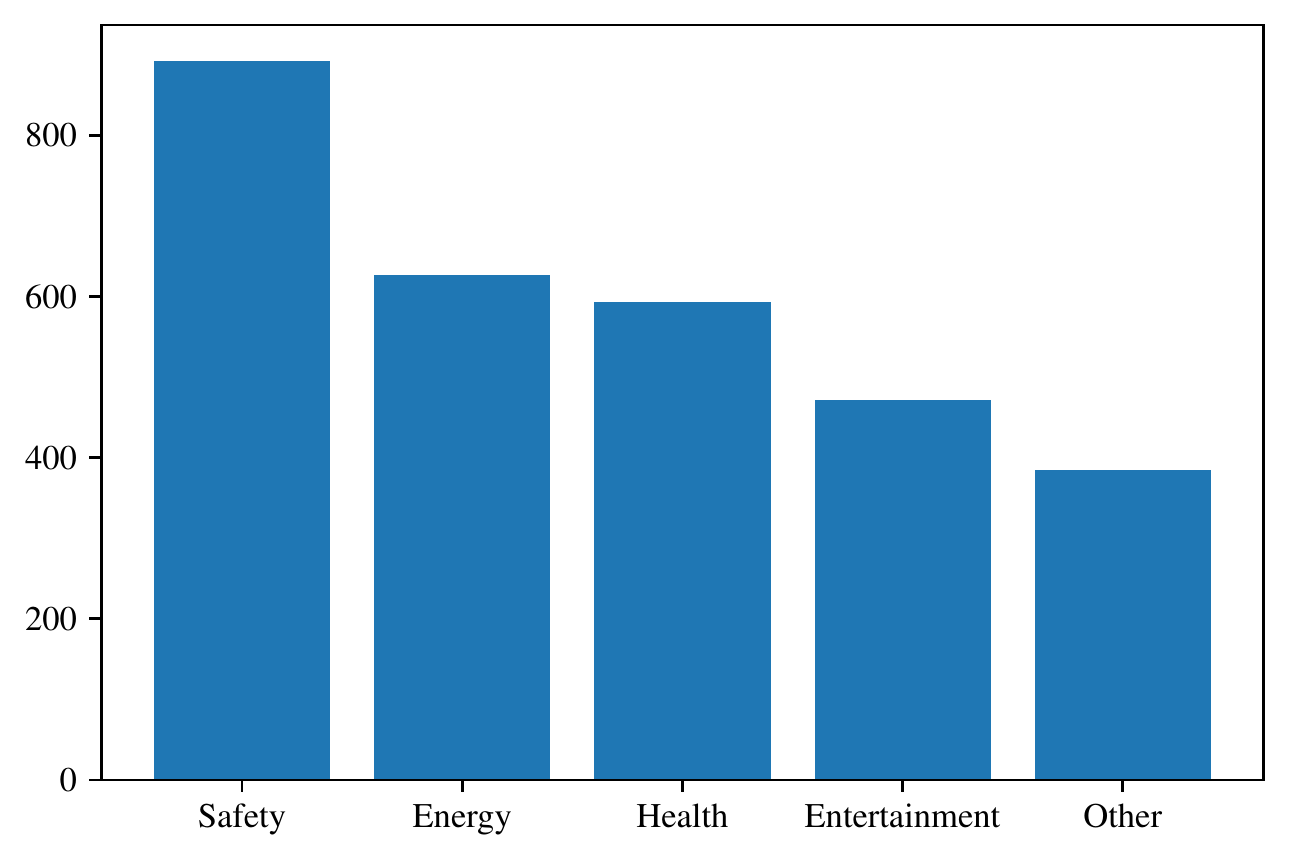}
    \caption{Distribution of domains within the user stories.}
    \label{fig:domain_analysis}
\end{figure}

Additionally, we need a measure to compare the proposed approaches. For this purpose, we used the domains assigned to the requirements in the corpus as labels. The domains are separated into five groups: \emph{Health}, \emph{Energy}, \emph{Entertainment}, \emph{Safety} and \emph{Other}. The \emph{Other} category contains additional user-defined, specific domains. For our study, we focus on the five top-level domains. \Cref{fig:domain_analysis} shows the distribution of domains within the user stories. The \emph{Safety} domain shows the most associated requirements while the least represented domain, \emph{Other}, exhibits roughly half the number of requirements. Still, the latter category contains about 400 requirements. Therefore, there is no under-representation of any of the categories. This labeling approach results in every requirement receiving a label and the number of categories remaining rather small, as to not over-complicate the supposed topics to be identified.

\subsection{Natural Language Processing Pipeline} 
\label{sub:own_pipeline}

\begin{figure}
  \centering
    \includegraphics[width=0.8\linewidth]{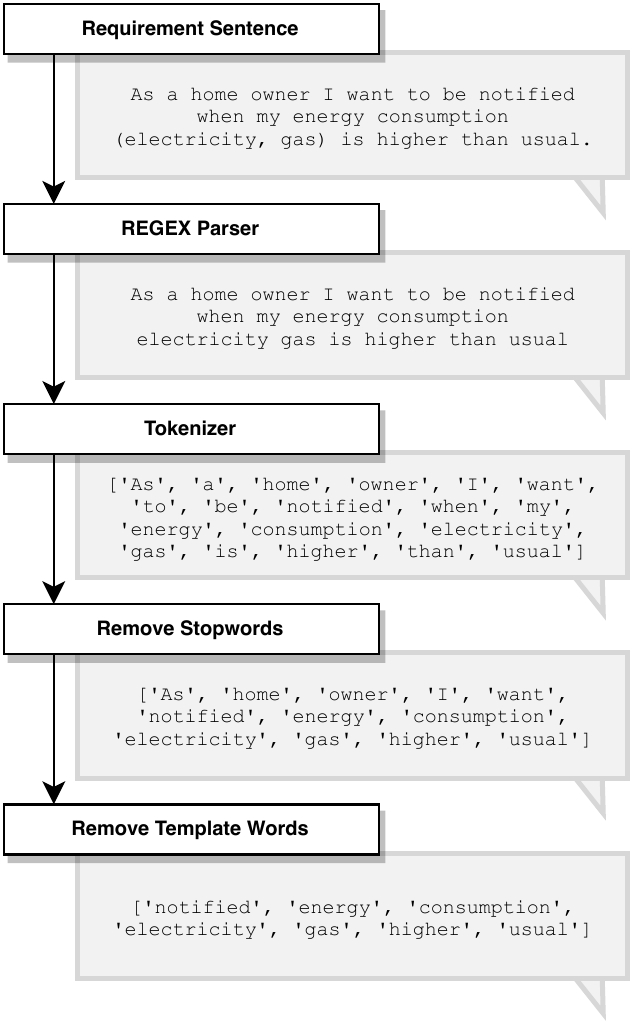}
    \caption{Processing an exemplary requirement sentence through our NLP preprocessing pipeline.}
    \label{fig:nlp_pipeline}
\end{figure}

In order to condition the data for further processing, we perform several Natural Language Processing (NLP) operations. Our NLP pipeline is shown in \Cref{fig:nlp_pipeline} with an exemplary application to a requirement from the dataset.

As some of the requirements sentences contain special characters, an initial data cleaning is necessary. We remove all but alphabetic characters as they do not provide any semantic value.

We apply \emph{tokenization} to separate the requirements into sequences of tokens~\cite{solangi_review_2018}. In our application, the tokens are simply the single words separated by whitespace. Therefore, all whitespaces and punctuation is removed from the requirements. The tokenization yields a list of tokens per requirement. After this step, the data exhibits 4,968 unique tokens.

\emph{Stopword-removal} is applied to remove words from the data that do not provide any semantic value~\cite{mhatre_dimensionality_2017}. In addition to the stopwords, we remove all template words\footnote{Predetermined template words in the \crowdre\ dataset: \textit{as}, \textit{smart}, \textit{home}, \textit{owner}, \textit{i}, \textit{want}, \textit{be}, \textit{able}.} from the requirements, as these are common to all requirements and therefore do not provide information to distinguish different topics. This reduces the size of the vocabulary, i.e. the number of unique tokens, to 4,851.


\begin{figure*}
    \centering
    \subfigure[Approach 1 \label{fig:approach_1_matrix}]
    {\includegraphics[scale=0.5]{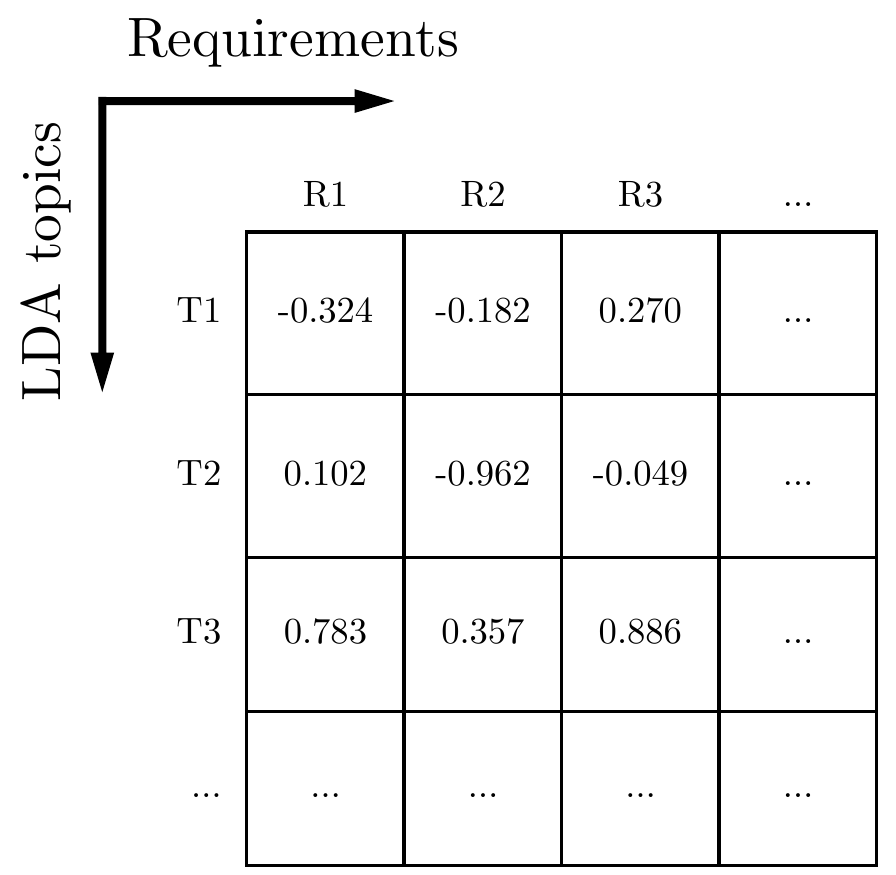}}
    \qquad
    \subfigure[Approach 2 \label{fig:approach_2_matrix}]
    {\includegraphics[scale=0.5]{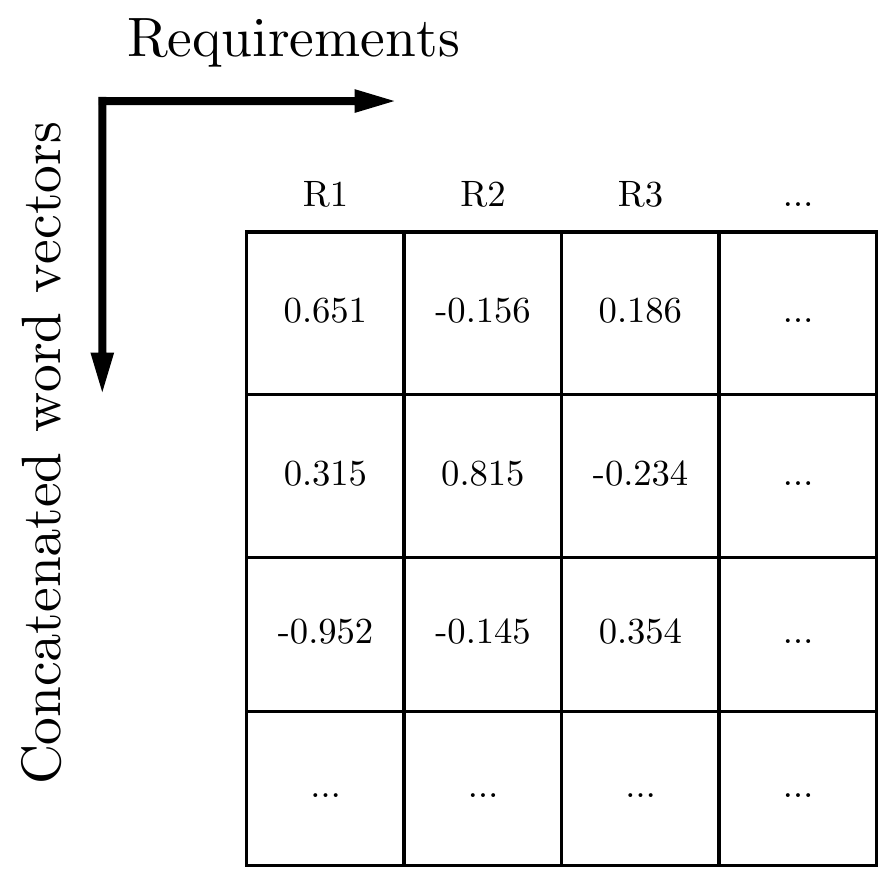}}
    \qquad
    \subfigure[Approach 3 \label{fig:approach_3_matrix}]
    {\includegraphics[scale=0.5]{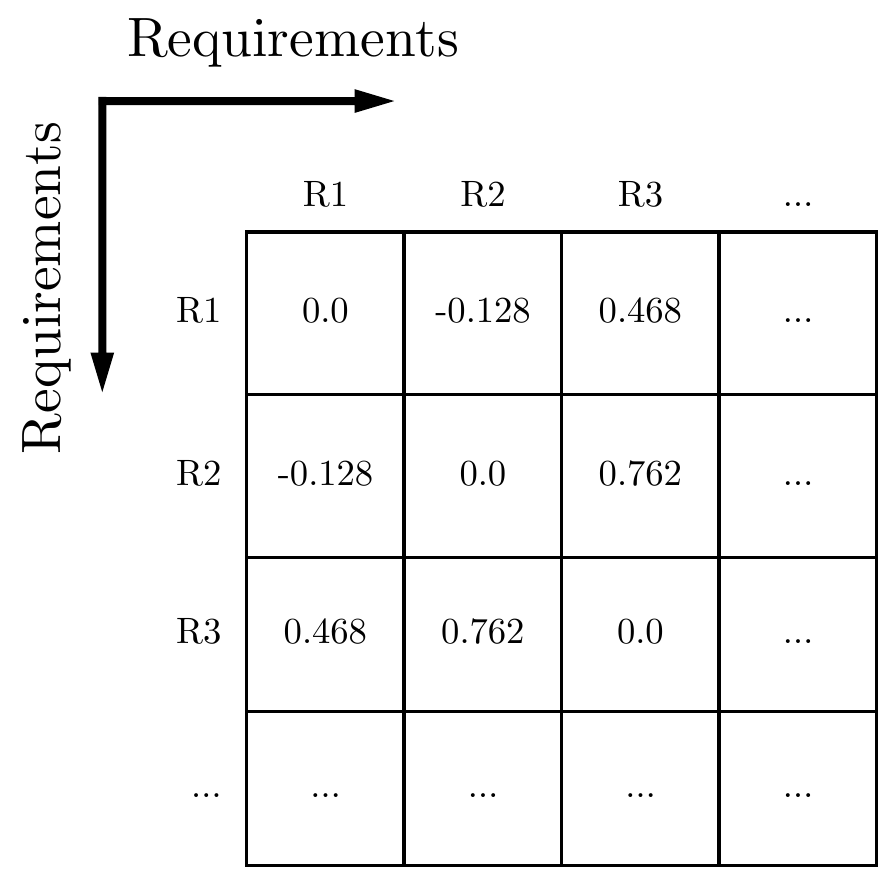}}
    \caption{Final representation of requirements for each approach before visualization.}
    \label{fig:embeddings_matrices}
\end{figure*}

\subsection{Approach 1: LDA} 
\label{sub:own_lda}

The presented LDA approach is mainly introduced to serve as a reference to the results of the two other neural network approaches. In \Cref{fig:approaches_overview}, this approach is denoted as \emph{A1}.

After initial preprocessing, \textit{Bag-of-Words} (BoW)~\cite{mikolov_efficient_2013} is applied to transfer the requirements to a numerical representation. Bag-of-Words is one of the basic techniques used to simplify sentences or documents in numerical space. In this application, a BoW vector is constructed for each requirement. Each vector has the size of the vocabulary of the dataset, i.e. the number of unique tokens in the dataset. For each requirement, there is a count per token in the vector representing how often each word in the vocabulary appears in the requirement.

Subsequently, a weighting scheme is applied to these BoW vectors, precisely we chose the Term Frequency -- Inverse Document Frequency (TF-IDF)~\cite{salton1988}. The term frequency is the rating how often a specific term occurs in the text, as already specified by the BoW vectors. The inverse document frequency is a measure of how relevant a term is in relation to all samples within the dataset~\cite{leskovec_data_2014}. For example, if a term occurs in every sample of the dataset, it is assumed to not be very informative towards differentiating samples. Very rarely occurring words are assumed to exhibit more explanatory power.

With every requirement represented as a weighted vector, an LDA is applied to identify the latent topics within the data. As our labeling approach from \Cref{sub:data_preparation} introduces five different labels to the requirements, we set the number of topics to be detected by the LDA to the same value. Therefore, the LDA produces vectors for the requirements containing the five probabilities that a requirement belongs to one of the identified topics. The resulting matrix is shown in \Cref{fig:approach_1_matrix}.

Lastly, we apply t-SNE~\cite{maaten_visualizing_2008}, a dimensionality reduction technique, to this matrix to get a 2-dimensional representation for each requirement, which can be plotted.

\subsection{Approach 2: Word Embeddings and PCA}
In order for our approach to include semantic aspects of the requirements, we apply word embeddings as introduced in \Cref{sub:back_word_embeddings}.
This approach is denoted in \Cref{fig:approaches_overview} as \emph{A2}.

For the second approach we compare two different implementations, one with self-trained word embeddings and one with pretrained ones. Our self-trained vectors are produced via the skip-gram method proposed by Mikolov~et~al.~\cite{mikolov_distributed_2013}. We construct 50-dimensional word vectors. This size is chosen due to the limited size of the vocabulary in the dataset. Most of the 4,851 words in the vocabulary only occur in the data very seldom. We empirically determine the best results to be obtained with a minimum word occurrence of five. i.e. all words that appear less than five times in the data, are dropped from the vocabulary and are not represented in the embedding. This results in about 24\,\% (1,159) of words being incorporated in the self-trained embedding.

With our dataset being relatively small and the created embedding not capturing all the semantic regularities due to the dropped words, we chose to also incorporate pretrained vectors. We use the word embedding from the Google News dataset\footnote{\url{https://code.google.com/archive/p/word2vec/}} trained on a set of about 100 billion words. Due to the large embedding and the extensive training data, the individual vectors exhibit 300 dimensions. We thus expect the quality of these word vectors to be much higher and, therefore, positively affect our topic modeling results, although we may loose the domain-specificity of self-trained embeddings.
Of the 4,851 unique tokens in our data, 93\,\% (4517 tokens) can be represented by the pretrained embedding. Tokens that are not included in the provided embedding are dropped. However, this loss in vocabulary only affects 13\,\% of the requirements, with the majority only missing one word.

To subsequently process the data, we first create a matrix for every requirement in the corpus by replacing each word with its vector representation. Due to the different lengths of the requirements, the resulting matrices have different dimensions. We apply a PCA to reduce the different dimensions to the length of the shortest requirement in the dataset, therefore producing requirements matrices of equal dimensions. For the approach with self-trained embeddings, the shortest requirement exhibits only one token. Therefore, after this dimensionality reduction, each requirement is represented as one 50-dimensional vector already. For the approach applying pretrained embeddings, the minimal requirement length is three tokens. Therefore, each requirement is represented by three 300-dimensional vectors in this case. We subsequently combine all these matrices to a single matrix $T$.

For the pretrained representation, the result is a 3-dimensional matrix $T\in\mathbb{R}^{n \times d \times s}$, where $n=2966$ is the total number of requirements, $d=300$ is the dimension of the word vectors and $s=3$ is the length of the shortest sample in the dataset. To be able to later plot the results, we concatenate all word vectors per requirement to receive one vector representing each requirement. The resulting matrix has dimensions $T' \in \mathbb{R}^{n \times d * s}$.

\Cref{fig:approach_2_matrix} shows the form of the resulting matrix for the approaches with self- and pretrained embeddings.
Finally, the matrices for each approach are processed via t-SNE to a reduce the dimensions per requirement for plotting.

\subsection{Approach 3: Word Mover's Distance} 
\label{sub:own_wmd}
As mentioned in \Cref{sub:word_movers_distance}, the document- or sentence-wise similarity cannot be captured by solely using word vectors. Therefore, the third and final approach employs word embeddings again but the subsequent processing is done with the \acf{WMD}. This approach is referred to in \Cref{fig:approaches_overview} as \emph{A3}.

As in the previous approach, we use and compare both the self-trained embedding as well as the pretrained one. We then apply the \ac{WMD} to calculate the distances between the requirements. The result is a distance matrix $D\in\mathbb{R}^{n\times n}$, with $n$ being the total number of requirements (see \Cref{fig:approach_3_matrix}). This matrix is subsequently reduced with t-SNE for plotting. The assumption is that requirements that are similar, show similar distances to all other requirements and are therefore plotted closely as well.


\section{Results} 
\label{sec:findings}

We expect to find four different topics in the dataset, one for each of the predefined application domains: \textit{Energy, Entertainment, Health}, and \textit{Safety}. We also assume sentences categorized as \textit{Other} to be visible as noise in the results, as these sentences may overlap topic-wise with the four concrete domains.
To visualize the outcome of each approach, we transform the word embedded user stories into 2-dimensional space using t-SNE. The marker colors and shapes follow the application domain the user stories are associated with: \emph{Health}~(purple\,$\circ$), \emph{Entertainment}~(beige\,$\times$), \emph{Energy}~(teal\,$\diamond$), \emph{Safety}~(cherry\,$+$), and \emph{Other}~(orange\,$\square$).

\subsection{LDA} 
\label{sub:findings_lda}

\begin{figure}
    \centering
    \includegraphics[width=\columnwidth]{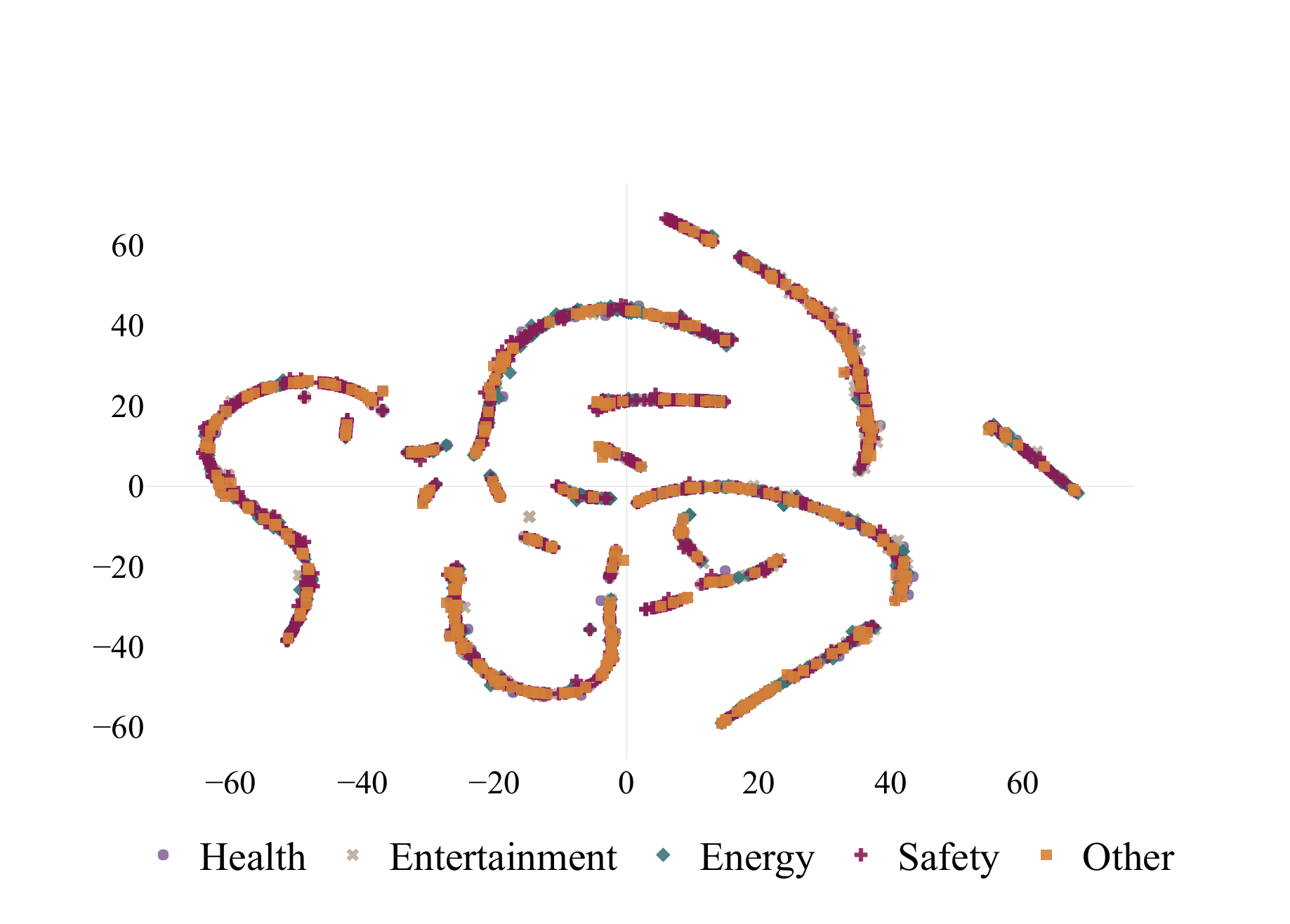}
    \caption{Results of approach A1: LDA with TF-IDF (plotted with t-SNE)}
    \label{fig:lda-tf-idf}
\end{figure}

\Cref{fig:lda-tf-idf} shows that the LDA approach results in separable clusters. Also, some similarities between the requirements plotted next to each other can be found. However, the clusters do not show a strong overlap with the original domain labels.


\subsection{Word Embeddings and PCA} 
\label{sub:findings_w2v}
\begin{figure*}
    \centering
    \subfigure[Self-trained word vectors\label{fig:w2v-selftrained}]{\includegraphics[width=\columnwidth]{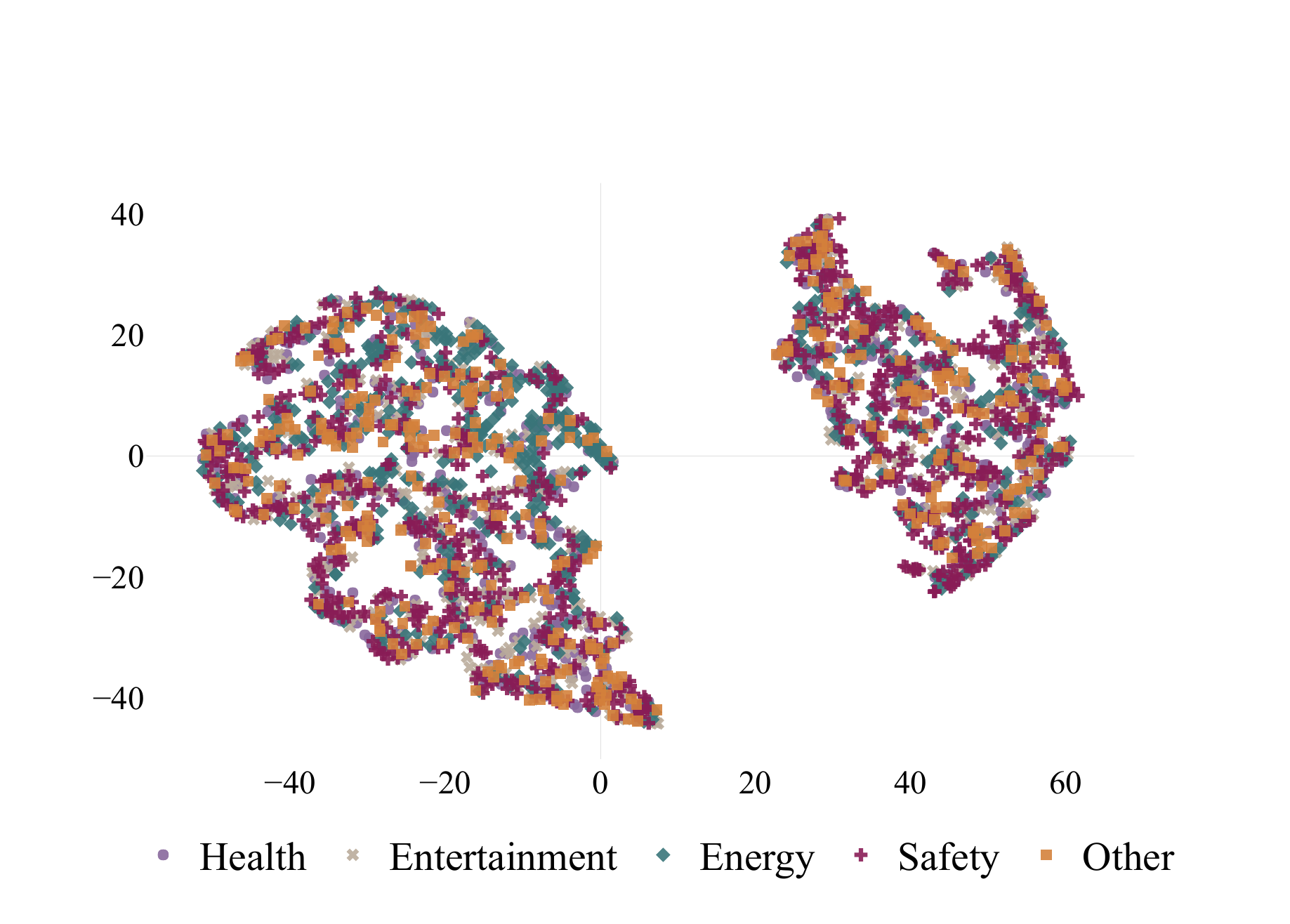}}
    \subfigure[Google News word vectors\label{fig:w2v-pretrained-4}]{\includegraphics[width=\columnwidth]{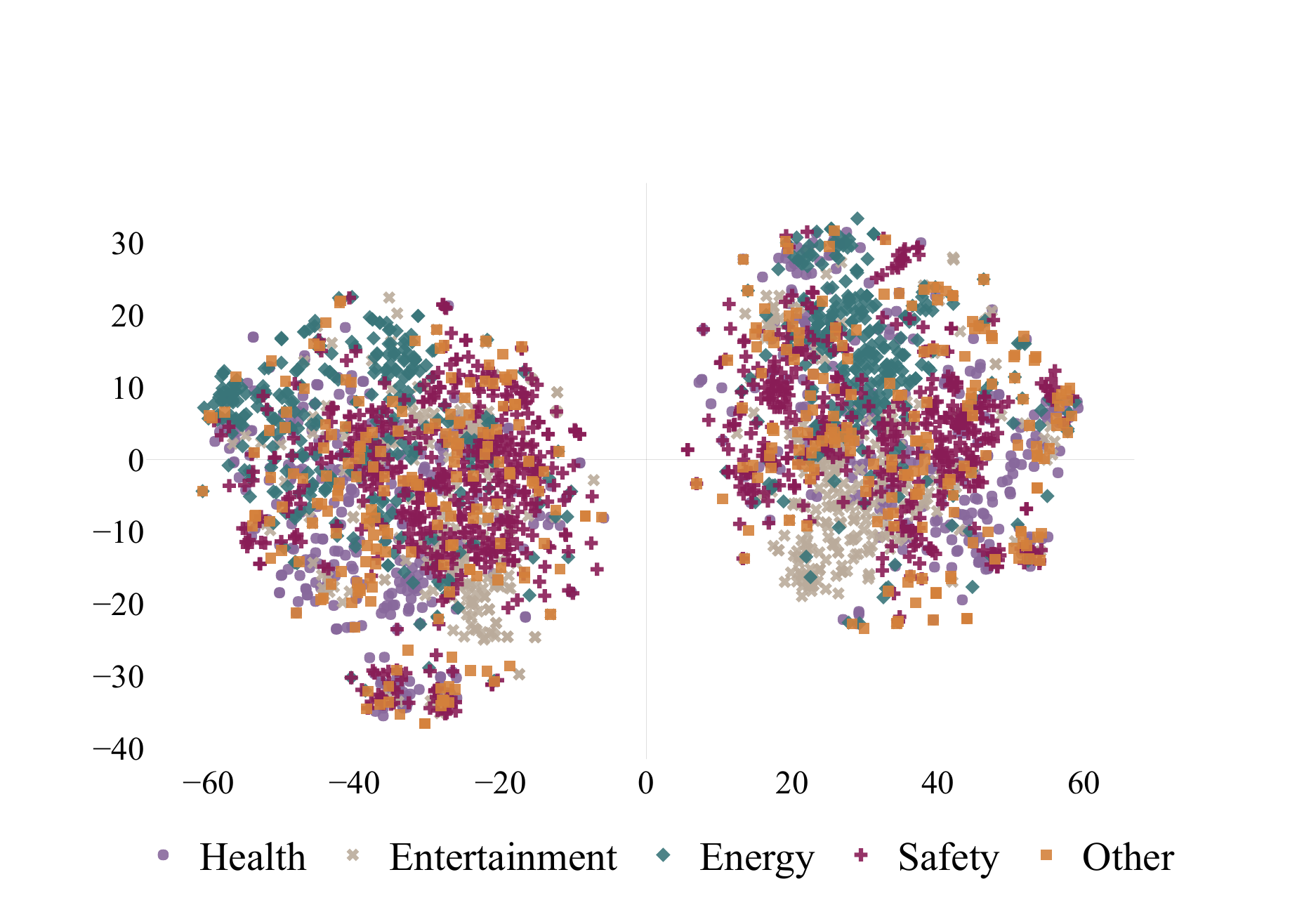}}
    \caption{Results of approach A2: Word embeddings and PCA (plotted with t-SNE)}
    \label{fig:w2v-pca}
\end{figure*}

As shown in \Cref{fig:w2v-pca}, we can identify two clusters in both plots resulting from the combination of word2vec and PCA for dimensionality reduction independent of the choice of self-trained (\Cref{fig:w2v-selftrained}) or pretrained (\Cref{fig:w2v-pretrained-4}) word embeddings. As with most machine learning techniques, it is difficult to say why exactly our approach resulted in these two clusters~\cite{ribeiro_why_2016}. Different settings for the perplexity and learning rate of the t-SNE do not change the number of clusters, at least. To better understand our results, we thus look into the plotted sentences and find the following:
\begin{enumerate}
  \item Sentences with multiple words in common are plotted close to each other.
  \item Sentences with fewer words in common are plotted further away from each other.
\end{enumerate}
As a consequence of (1), we achieve good results for requirements that overlap in vocabulary (e.g.\ \textit{``As a home owner I want Room thermostat sensor so that The room is optimal temperature for an occupant''} at (-58.033, 11.973) and \textit{``As a home occupant I want Room thermostats so that Protect the room temperature''} at (-57.890, 12.276)). However, because of (2), sentences that express related requirements in different words are not clustered reliably. E.g.\ \textit{``As a home occupant I want music to be played when I get home so that it will help me relax''} at (-24.767, 4.210) and \textit{``As a home owner I want music to play whenever I am in the kitchen so that I can be entertained while cooking or cleaning''} at (53.929, 4.752). We anticipated the latter findings due to the shortcomings of word2vec to identify similar sentences as mentioned in \Cref{sub:word_movers_distance}. Therefore, we cannot model the topics as desired. Nevertheless, this approach delivers deeper insight into the dataset and needs relatively little computation time, as the results are available within a few minutes.

\subsection{Word Embeddings and Word Mover's Distance} 
\label{sub:findings_wmd}
\begin{figure*}
	\centering
	\subfigure[Self-trained word vectors\label{fig:wmd-selftrained-1}]{\includegraphics[width=\columnwidth]{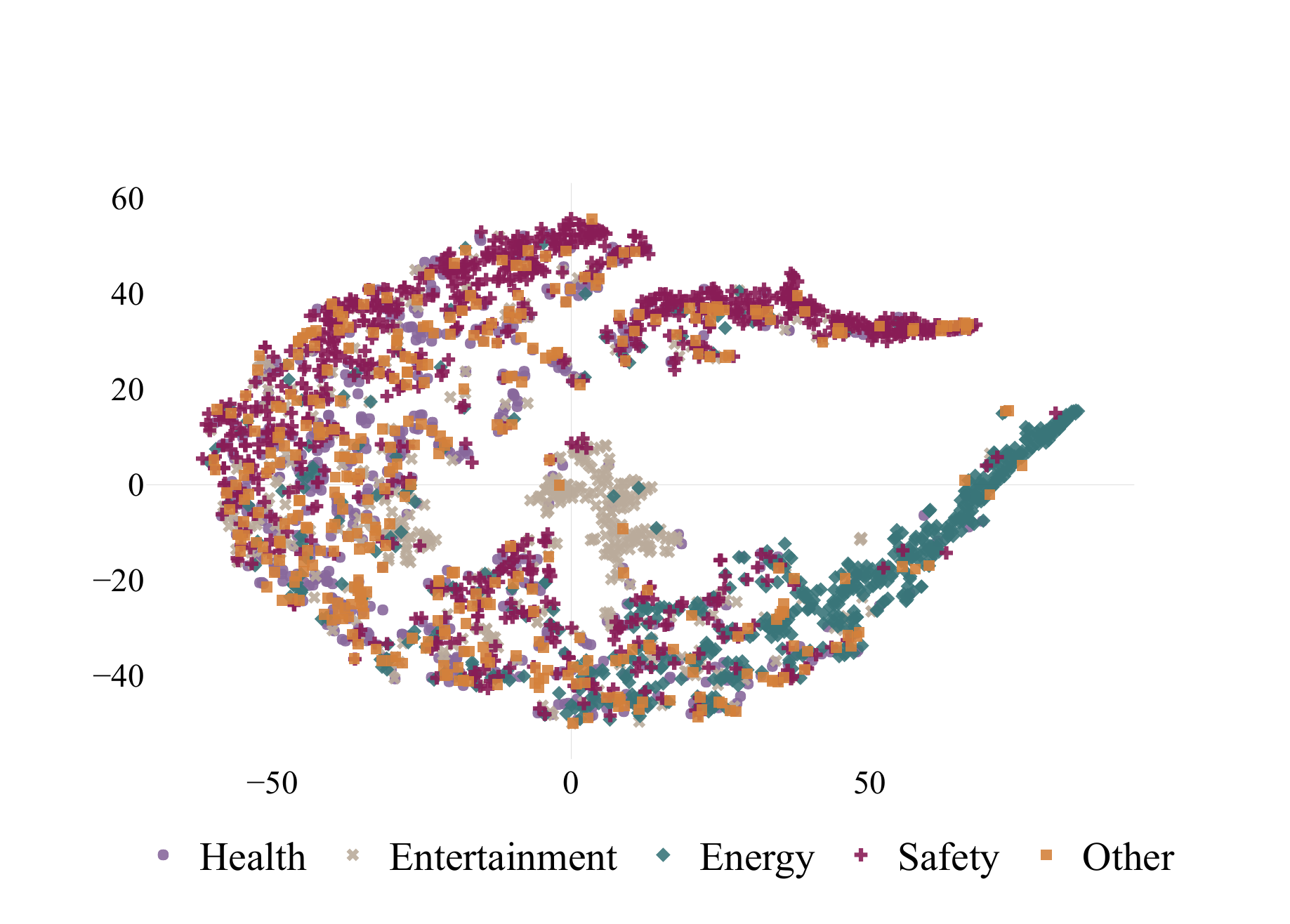}}
    \subfigure[Google News word vectors\label{fig:wmd-pretrained-1}]{\includegraphics[width=\columnwidth]{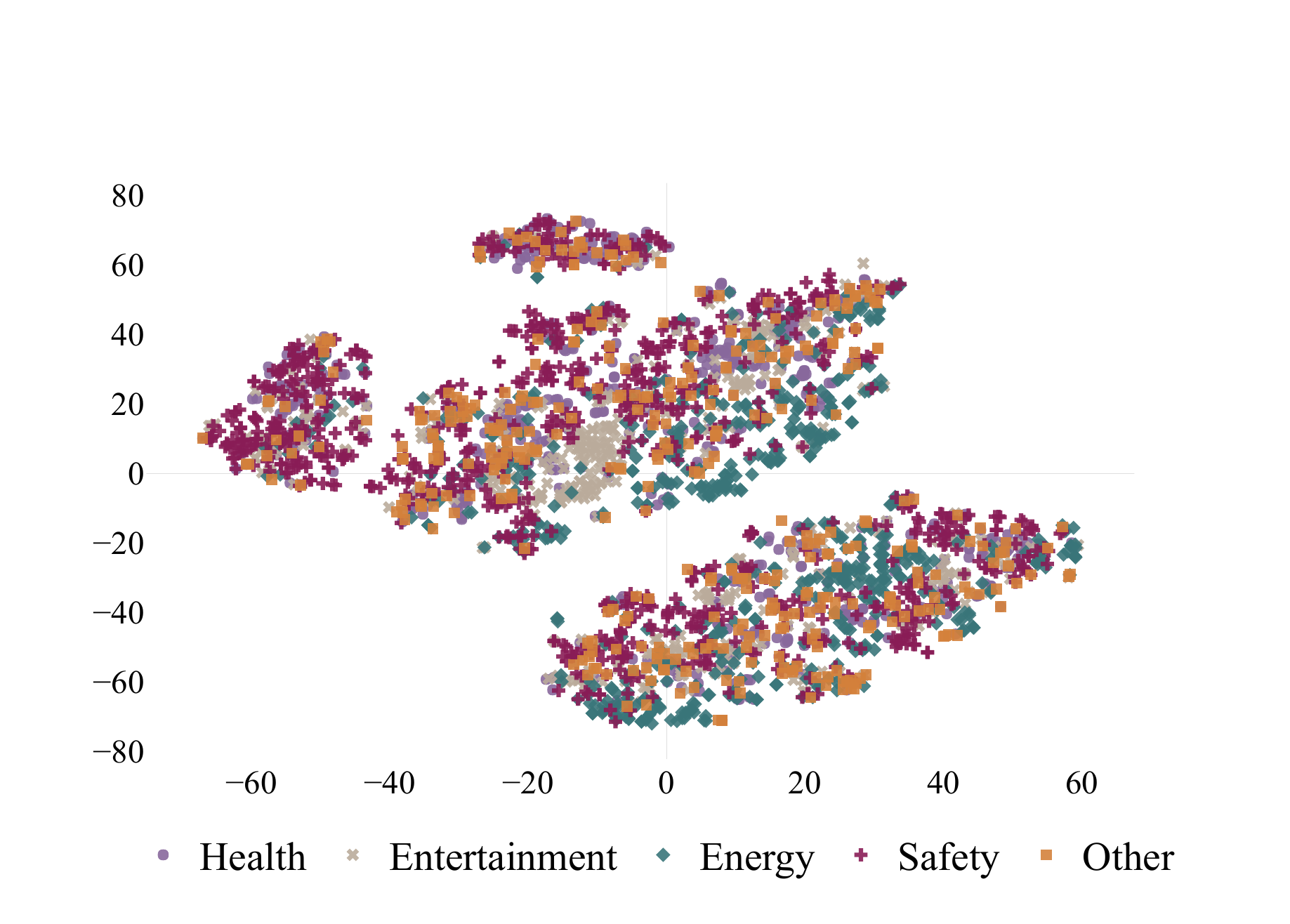}}
    \caption{Results of approach A3: Word embeddings and Word Mover's Distance (plotted with t-SNE)}
    \label{fig:wmd-comparison}
\end{figure*}

We achieve the best results in our third approach, using word2vec and WMD. \Cref{fig:wmd-comparison} shows the plotted distance matrices we created, as described in \Cref{sub:word_movers_distance}. Here, we can successfully distinguish clusters both spatially and content-wise. Using the self-trained word vectors, in \Cref{fig:wmd-selftrained-1} we can see that the domains \textit{Entertainment} (gray sentences around (0,0)) and \textit{Energy} (stretching from (0,-45) to (10,57)) can be distinguished clearly. Also, a cluster predominantly consisting of \textit{Health} requirements is apparent in the region from (0,20) to (70,45).

Judged by the domain categories only, the clustering with our self-trained word vectors seems to yield better results. But as manual inspection shows, the clustering based on the Google News vectors also brings new insights into the dataset: In \Cref{fig:wmd-pretrained-1}, we can see a much clearer demarcation between the clusters. Also, two new domains become apparent. Although the sentences seem unrelated at first (told by the different label colors), the leftmost cluster, namely the area between \mbox{(-65,-5)} and \mbox{(-40,40)}, mostly contains sentences related to parenting and children. Furthermore, the topmost cluster between (\mbox{-30,55}) and \mbox{(0,70)} contains requirement sentences about animals. These results show that the dataset may be clustered into different clusters than the 4 domain-based clusters we initially anticipated.



\section{Discussion}
\label{sec:discussion}
\subsection{Limitations and Threats to Validity}
The \crowdre dataset contains requirements in the form of user stories. We assume that the structured form of user stories may facilitate any form of automated analysis (see~\cite{Dalpiaz19,Lucassen17}). Although the formulations within the free-text parts of the user stories are quite different in terms of length and used words, we cannot say how the approaches would work when applied to unrestricted natural language requirements.

Even though the \crowdre{} dataset is too large to process the requirements manually, it is relatively small for the application of automatic topic modeling techniques: LDA is a technique proven to work well on large documents. Short texts instead, contain very limited word co-occurrence information. This hinders the LDA to work well on short texts~\cite{quan_short_2015}, as we have also seen in our results.

The word2vec approach is impacted by the text length as well, since the document similarity cannot be accurately measured under BoW representations due to the extreme sparseness of short texts.~\cite{li_classifying_2019}. Also, when working with word embeddings, in general more data (as opposed to simply relevant data) creates better embeddings~\cite{kusner_word_2015}. As already mentioned, to benchmark word2vec, Mikolov~et~al.\ trained their tool on the Google News dataset with 100 billion words, so a dataset 2000 times the size of our dataset. This suggests, better results may be possible using the same techniques on a larger data set. However, the meaning of requirements usually depends on the considered application domain~\cite{ferrari_natural_2018}. Therefore, domain-specific word embeddings may lead to better results~\cite{Ferrari19}. In our case, however, we achieved the best results using a pretrained general-purpose model. This may indicate that the advantages of domain-specific word embeddings are overruled by the disadvantages of the small dataset. For the future, it may be interesting to use domain-specific word embeddings trained on larger data sets (e.g.\ Wikipedia or news paper articles on home automation). 

All of our approaches associate a user story with a point in a high-dimensional vector space. We applied dimensionality reduction techniques (PCA and t-SNE) to be able to compare the results of the approaches visually. Dimensionality reduction techniques provide an approximation of the original data, which may result in information being lost in the process~\cite{wold_principal_1987}. A more precise analysis of clusters may be possible by clustering the points directly in the high-dimensional space (e.g.\ by applying \emph{k-means}). We did that for some of our results and found that the clusters generated by k-means are similar to the visually distinguishable clusters in the 2-dimensional plots.   


For a proper and quantified evaluation of our results, manual work would be needed. To rate our findings, the dataset has to be labeled properly. Consider the following example:

\begin{enumerate}
	\item[RE1] \textit{``As a home occupant I want music to be played when I get home so that it will help me relax''} (\emph{Health})
	\item[RE2] \textit{``As a home owner I want music to play whenever I am in the kitchen so that I can be entertained while cooking or cleaning''} (\emph{Energy})
\end{enumerate}
 
With our word2vec \& WMD approach, these sentences are plotted nearby, both located inside the central \emph{Entertainment} cluster in \Cref{fig:wmd-selftrained-1}. We cannot say that RE1 is surely assigned to the wrong domain, we consider a relationship to the \emph{Entertainment} cluster to be equally valid, though. Also, with RE2 the \emph{Energy} domain may have been selected accidentally, as the domains are next to each other in the select box of the form the crowd workers used when they created the requirements. When manually reviewing the dataset to fix the labels, our results could also be improved through generally cleaning the dataset. Cleaned datasets have a much higher impact on the training results of ML models than the optimization of hyperparameters~\cite{chu_data_2016,krishnan_data_2016}.

\subsection{Future Work}

Besides cleaning the data, future work can be done for cross-validation and performance improvements: Li~et~al.\ also created a classifier using WMD~\cite{li_classifying_2019}. Using their approach one could create clusters on the \crowdre{} dataset to compare the findings with our results. Regarding performance improvements, the calculation of the WMD matrix is relatively time-consuming. Wu~et~al.\ propose a different distance measure for document clustering, which, compared to the WMD, \textit{``can achieve much lower time complexity with the same accuracy''}~\cite{wu_topic_2017}.
In a new approach called Word Mover's Embeddings, Wu~et~al.\ also use pretrained word embeddings and were able to improve the accuracy and the calculation effort when they tested the approach on several benchmark text classification datasets~\cite{wu_etal_2018_word}.

Finally, our work may be used in future attempts to crowd source user requirements for input validation in a web service. When continuously learning and storing the word vectors for new requirements, it would be possible to already suggest similar sentences to the ones a crowd worker is about to enter, based on WMD. If the crowd worker obtains that his submission overlaps with an existing sentence, they could up-vote the existing sentence instead of submitting their sentence. This would not only avoid duplication, but also help in data-driven RE to identify frequently requested requirements without the need for additional data processing.


\section{Conclusion} 
\label{sec:conclusion}
Acquiring requirements and requirements-related information from crowd workers facilitates a user-centered RE process and enables engineers to consider requirements form a broad and heterogeneous set of potential users~\cite{Groen17}. However, crowd-sourced information or information from other user feedback platforms are raw and unstructured. Automatic techniques are essential to preprocess, filter, and analyze the large amount of gathered information. In this paper, we have proposed and compared three approaches for clustering crowd-sourced requirements given in the form of user stories. A ``classical'' approach based on Latent Dirichlet Allocation and two approaches based on similarity measures in vector space models generated from different word embeddings. To the best of our knowledge, a combination of word embeddings with Word Mover's Distance as distance measure has not been used for requirements clustering.

Our main reference for evaluation was a mapping of user stories to one of five domains, which was defined by the authors of the user stories. In our evaluation, a combination of a vector space model based on a pretrained word embedding (\emph{word2vec}) and \ac{WMD} as distance measure resulted in the most interesting results. Most interesting means that the approach resulted in a reasonable number of clusters with good overlap to the original domains. In some sample cases, the clustering pointed to potential misclassifications by the authors. 

\bibliographystyle{IEEEtran}
\bibliography{references}

\end{document}